\documentclass[]{ceurart}

\usepackage{url}
\usepackage{graphicx}
\usepackage{multirow}
\usepackage{placeins}
\usepackage{adjustbox}
\usepackage{caption}
\usepackage{subcaption}

\begin{document}

\copyrightyear{2023}
\copyrightclause{Copyright for this paper by its authors.
  Use permitted under Creative Commons License Attribution 4.0 International (CC BY 4.0).}

\conference{KBC-LM'23: Knowledge Base Construction from Pre-trained Language Models workshop at ISWC 2023}

\title{Using Large Language Models for Knowledge Engineering (LLMKE): A Case Study on Wikidata}

\author[1]{Bohui Zhang}[%
email=bohui.zhang@kcl.ac.uk,
url=https://bohuizhang.github.io/,
]
\address[1]{Department of Informatics, King's College London, London, UK}

\author[1]{Ioannis Reklos}[%
email=ioannis.reklos@kcl.ac.uk,
]

\author[1]{Nitisha Jain}[%
email=nitisha.jain@kcl.ac.uk,
url=https://nitishajain.github.io/,
]

\author[1]{Albert Mero\~{n}o Pe\~{n}uela}[%
email=albert.merono@kcl.ac.uk,
url=https://www.albertmeronyo.org/,
]

\author[1]{Elena Simperl}[%
email=elena.simperl@kcl.ac.uk,
url=http://elenasimperl.eu/,
]

\begin{abstract}
In this work, we explore the use of Large Language Models (LLMs) for knowledge engineering tasks in the context of the ISWC 2023 LM-KBC Challenge. For this task, given subject and relation pairs sourced from Wikidata, we utilize pre-trained LLMs to produce the relevant objects in string format and link them to their respective Wikidata QIDs. We developed a pipeline using LLMs for Knowledge Engineering (LLMKE), combining knowledge probing and Wikidata entity mapping. The method achieved a macro-averaged F1-score of 0.701 across the properties, with the scores varying from 1.00 to 0.328. These results demonstrate that the knowledge of LLMs varies significantly depending on the domain and that further experimentation is required to determine the circumstances under which LLMs can be used for automatic Knowledge Base (e.g., Wikidata) completion and correction. The investigation of the results also suggests the promising contribution of LLMs in collaborative knowledge engineering. LLMKE won Track 2 of the challenge. The implementation is available at: \url{https://github.com/bohuizhang/LLMKE}. 
\end{abstract}

\maketitle

\section{Introduction}\label{introduction}
Language models have been shown to be successful for a number of Natural Language Processing (NLP) tasks, such as text classification, sentiment analysis, named entity recognition, and entailment. The performance of language models has seen a remarkable improvement since the advent of several LLMs such as ChatGPT\footnote{https://chat.openai.com/} and GPT-4~\cite{openai-2023-gpt-4} models from OpenAI, LLaMa-1~\cite{touvron-et-al-2023-llama} and Llama 2~\cite{touvron-et-al-2023-llama-2} from Meta, Claude\footnote{https://claude.ai/} from Anthropic, and Bard\footnote{https://bard.google.com/} from Alphabet.

This surge in the development and release of LLMs, many of which have been trained with Reinforcement Learning with Human Feedback (RLHF), has allowed users to consider the LMs as \textit{knowledge repositories}, where they can interact with the models in the form of `chat' or natural language inputs. This form of interaction, combined with the unprecedented performance of these models across NLP tasks, has shifted the focus to the engineering of the input, or the `prompt' to the model in order to elicit the correct answer. Subsequently, there has been a steady increase in research outputs focusing on prompt engineering in the recent past~\cite{shin-et-al-2020-autoprompt,qin-eisner-2021,liu-et-al-2023-survey}.

Knowledge graphs (KGs) are a technology for knowledge representation and reasoning, effectively transferring human intelligence into symbolic knowledge that machines can comprehend and process~\cite{ehrlinger-wob-2016, fensel-et-al-2020-kg, hogan-et-al-2021-kg}. The process of creating these KGs, referred to as knowledge engineering, is not trivial, either automatically or collaboratively within human communities~\cite{simsek-et-al-2022}. Wikidata~\cite{vrandecic-krotzsch-2014-wikidata}, as the largest open KGs, contains rich knowledge of real-world entities. It has been developed in a collaborative manner, with contributions from a community of users and editors~\cite{piscopo-simperl-2018}.

While the concept of using LMs to construct and complete KGs has been extensively explored in previous research~\cite{razniewski-et-al-2021,alivanistos-et-al-2022-prop,veseli-et-al-2023}, the recent surge in LLMs performance has rekindled discussions about the possibility of leveraging the strengths of both technologies and unifying them~\cite{pan-et-al-2023-llm-kg}.
Despite the immense potential offered by LLMs as knowledge bases, there exist fundamental disparities that differentiate them from KGs. The most pivotal of these distinctions lies in the domain of reasoning. Not only do traditional KGs store facts, they also impose logical constraints on the entities and relations in terms of defining the types of the entities as well as prescribing the domain and range of the relations. The capability of LLMs for logical reasoning remains unclear and appears to face challenges~\cite{wei-et-al-2022-cot,huang-chang-2023}.
Moreover, the most widely adopted and successful LLMs have been trained on data obtained from publicly available sources, and due to the inherent limitations of the training method of these models, they tend to exhibit expert-level knowledge in popular domains or entities while often displaying a limited understanding of lesser-known ones.

In this paper, we describe our approach LLMKE to using LLMs for Knowledge Engineering tasks, especially targeting solving the ISWC 2023 LM-KBC Challenge~\cite{singhania-et-al-2023-lm-kbc}, and report our findings regarding the prospect of using these models to improve the efficiency of knowledge engineering. The task set by this challenge is to predict the object entities (zero or more) given the subject entity and the relation that is sourced from Wikidata. For instance, given the subject \textit{Robert Bosch LLC} with Wikidata QID \texttt{Q28973218} and the property \textit{CompanyHasParentOrganisation}, the task is to predict the list of object(s), [`\textit{Robert Bosch}'] and their matched QID(s), [`\texttt{Q234021}']. We used two state-of-the-art LLMs, gpt-3.5-turbo\footnote{https://platform.openai.com/docs/models/gpt-3-5} and GPT-4 for this task. By performing different experiments using in-context learning approaches, we have been able to achieve a macro-average F1 score of 0.701, with F1-scores ranging from 0.3282 in the \textit{PersonHasEmployer} property to 1.0 in the \textit{PersonHasNobelPrize} property.

\section{Related Works}\label{related-works}
\subsection{LLMs for Knowledge Probing}

The ability of LLMs to perform knowledge-intensive tasks, especially knowledge probing, has been extensively investigated~\cite{zhong-et-al-2021-optiprompt,liang-et-al-2022-helm,peng-et-al-2022-copen}.
In particular, several previous works have attempted to use language models to construct or complete KGs. Among early works, the LAMA paper by Petroni et al.~\cite{petroni-et-al-2019-lama} investigated the task of knowledge graph completion by probing LMs to extract facts via cloze-style prompts. Along similar lines, KG-BERT leverages the BERT language model to perform the link prediction task for knowledge graph completion\cite{yao-et-al-2019-kg-bert}.
The extent of the usefulness of LLMs for the construction and completion of knowledge graphs has since been further analyzed~\cite{razniewski-et-al-2021}. Follow up work after LAMA improved the performance even further~\cite{qin-eisner-2021,zhong-et-al-2021-optiprompt}. 
Recently, Veseli et al.~\cite{veseli-et-al-2023} have performed a systematic analysis on the potential of LMs for automated KG completion. They report that LMs can be useful for predicting facts with high precision for some relations in Wikidata, though this is not generalizable.
Prompt engineering has caught the attention of many recent works that aim to elicit knowledge from the language models~\cite{alivanistos-et-al-2022-prop}. These works are the most similar to our approach in this paper.

\subsection{Knowledge Probing Benchmarks}

To fulfil the need for comprehensively investigating the ability of LLMs to perform knowledge-intensive tasks, there has been a growing trend of knowledge-oriented benchmarks and datasets. These benchmarks encompass diverse domains, address various scenarios, including question answering, reading comprehension, and fact completion, and represent knowledge in different formats, including queries, cloze-style, incomplete triples, etc~\cite{liang-et-al-2022-helm,rogers-et-al-2023-qa-taxonomy}. And knowledge graphs, especially the large-scale and general-purpose ones, have become vital sources for constructing these benchmarks. As the pioneering dataset in the language models era, LAMA was constructed from a variety of knowledge graph sources of factual and commonsense knowledge, including T-REx~\cite{elsahar-et-al-2018-t-rex}, ConceptNet~\cite{speer-havasi-2012-conceptnet}, etc. There are several benchmarks that evolved from it to overcome its limitations and expand its abilities, such as KAMEL~\cite{kalo-fichtel-2022-kamel} which extended LAMA from single-token objects to multi-token ones. KILT~\cite{petroni-et-al-2021-kilt} was constructed from millions of Wikipedia pages spanning a wide range of knowledge-intensive language tasks. WikiFact~\cite{liang-et-al-2022-helm} as a part of the HELM benchmark is the most similar to this challenge, where they use Wikidata relations and triples to construct the benchmark. But the challenge used a different evaluation paradigm. KoLA~\cite{yu-et-al-2023-kola} aimed at measuring the real-world performance of LLMs by expanding beyond language modeling, adding evolving data sources, and attempting to measure the ability of the models in all facets of knowledge processing, ranging from knowledge memorization to knowledge creation. The data sources it used are also highly overlapping with Wikidata and Wikipedia.

\section{Methods}\label{methods}
\subsection{Problem Formulation}

Most of the previous works on using LLMs for fact completion stop at the string level, which leaves gaps for constructing hands-on knowledge graphs and thus hinders downstream application. Our work pushed a step forward on this task, where the extracted knowledge is not only in string format but also linked to their respective Wikidata entities. Formally, given a query consisting of subject entity $s$ and relation $r$, the task is to predict a set of objects $\{o_i\}$ with unknown numbers ($\vert\{o_i\}\vert \geq 0$) by prompting LLMs and mapping the objects to their related Wikidata entities $\{w_{o_i}, \cdots, w_{o_n}\}$.

\subsection{The LLMKE Pipeline}\label{llmke}

\subsubsection{Knowledge Probing}

The pipeline consists of two steps: \textit{knowledge probing} and \textit{Wikidata entity mapping}. For the knowledge probing step, we engineered prompt templates for probing knowledge from LLMs. We adopt OpenAI's gpt-3.5-turbo and GPT-4 in this step. For each of the LLMs, we run experiments with three types of settings. The first is question prompting, where LLMs are provided with questions as queries. For example, ``\textit{Which countries share borders with Brazil?}". The second is triple completion prompting, where prompts are formatted as incomplete triples, such as ``\textit{River Thames, RiverBasinsCountry:}''. There are several heuristics employed in these two settings. For example, there are only 5 different Nobel Prizes, so \textit{PersonHasNobelPrize} has 6 candidate answers, including the empty answer. When the answer space is limited, providing all potential answers in the prompt templates is likely to reduce the difficulty of formatting and disambiguating the objects, thus helping LLMs perform well.

In the third setting, we provide retrieval-augmented context to help LLMs by enriching knowledge from corpus, including Wikipedia and domain-specific websites. Trying to leave space for invoking the `critical thinking' of LMs and for further investigating the effect of adding context, the prompts used in this setting are separated into two steps. At first, we ask LLMs to predict the objects based on their own knowledge using the same settings as question prompting. In the second step, we provided the context knowledge, and LLMs were asked to make predictions again by considering the context and comparing it with the previous response. The prompt is like `\textit{Given the context: [retrieval-augmented context], compared and combined with the previous predictions, [question prompt]}'. In this case, we let LLMs to decide whether they will insist on their own knowledge or change their answers based on the context. In this study, we used Wikipedia as the general-domain context source. The first paragraphs of the entity's Wikipedia page (the introduction) and the JSON format of the Wikipedia Infobox are organized and provided to LLMs. For relations that could potentially have empty results, the prompt indicated the required return format (i.e., [""]).

In all settings, we perform few-shot learning, where we provide three examples (i.e., prompt and answer pairs) from the training set. Since the required format of results is a list, providing examples with the exact format is expected to help LLMs return better-formatted results.

\subsubsection{Wikidata Entity Mapping}\label{disambiguation}

The entity mapping step first finds Wikidata entities for each object string using the MediaWiki Action API\footnote{https://www.wikidata.org/w/api.php}. One of the actions, \textit{wbsearchentities}\footnote{https://www.wikidata.org/w/api.php?action=help\&modules=wbsearchentities} which searches for entities using labels and aliases, returns all possible Wikidata entities as candidates. Then, in the disambiguation step, the actual Wikidata entities linked to the objects are selected. The \textbf{baseline} disambiguation method selects the first entity from the list of candidates returned by the \textit{wbsearchentities} action, which is notably incorrect. To reduce the cost while improving the accuracy for disambiguation, we treated different relations with three \textbf{improved} methods: \emph{case-based}, \emph{keyword-based}, and \emph{LM-based}. 

The \emph{case-based} method is a hard-coding solution for efficiently solving ambiguities for relations with smaller answer spaces and limited corner cases. It is built on the baseline method by adding the function that maps specific objects to their respective Wikidata QIDs. For example, \textit{CompoundHasParts} only has all the chemical elements as its answer space. Further, there is only one mistake in the baseline method, `\textit{mercury}'. Thus, when predicting for \textit{CompoundHasParts}, the case-based method always maps `mercury' in the object lists to \texttt{Q925} (the chemical element with symbol Hg) instead of \texttt{Q308} (the planet). For other relations with a larger answer space but also entities with common characteristics, we used the \emph{keyword-based} method, which extracts the description of the candidate entities from its Wikidata page and searches entities with their description using relevant keywords. This method is used when there are common words in the entity description. For example, object entities of the relation \textit{CountryHasOfficialLanguage} always have the keyword `language' in their descriptions.

The above two methods clearly suffer from limitations due to their poor coverage and inflexibility. The third method is language model-based (\emph{LM-based}). We constructed a dictionary of all candidate QIDs with their labels as keys and descriptions as values, concatenated it with the query in this first step, and asked LMs to determine which one should be selected. This method is used when there is no semantic commonality between the answers and disambiguation is required to understand the difference between entities, e.g., properties with the whole range of human beings as potential answers such as `\textit{PersonHasSpouse}'. As there is no commonality among the labels and descriptions of answers, the decision is left to the LMs. This method also has limitations, such as being time-consuming and unstable.

\section{Results}\label{results}
\subsection{Datasets}

The dataset used in the ISWC 2023 LM-KBC Challenge~\cite{singhania-et-al-2023-lm-kbc} is queried from Wikidata and further processed. It comprises 21 Wikidata relation types that cover 7 domains, including music, television series, sports, geography, chemistry, business, administrative divisions, and public figure information. It has 1,940 statements for each train, validation, and test sets. The results reported are based on the test set.\footnote{To investigate the actual knowledge gap between LLMs and Wikidata, we created ground truths of the test set through Wikidata SPARQL queries for offline evaluation. We report and analyze the offline evaluation results in Section~\ref{results} and the online evaluation results from CodaLab in Appendix~\ref{online}.} In the dataset, the minimum and maximum number of object-entities for each relation is different, ranging from 0 to 20. The minimum number of 0 means the subject-entities for some relations can have zero valid object-entities, for example, people still alive should not have a place or cause of death.

\subsection{Model Performance}

In terms of the overall performance of the model as shown in Table~\ref{tab:model-performance} and~\ref{tab:online-evaluation}, GPT-4 is better than gpt-3.5-turbo. The retrieval-augmented context setting has the best performance compared with the other two few-shot learning settings. And the performance on question answering prompts and triple completion prompts is quite close. 

\begin{table}[ht]
    \centering
    \caption{Comparison of the performance of gpt-3.5-turbo and GPT-4 models based on the three settings: \textbf{question} prompting, \textbf{triple} completion prompting, and retrieval-augmented \textbf{context} setting. `baseline' and `improved' represent different disambiguation methods documented in Section~\ref{disambiguation}. The best F1-scores among the three settings and two disambiguation methods of the models are highlighted.}
    \label{tab:model-performance}
    \begin{adjustbox}{width=1\textwidth}
    \begin{tabular}{cc|ccc|ccc|ccc}
    \hline
        \multirow{2}{*}{\textbf{Model}} & \multirow{2}{*}{\textbf{Disambiguation}} & \multicolumn{3}{c}{\textbf{question}} & \multicolumn{3}{c}{\textbf{triple}} & \multicolumn{3}{c}{\textbf{context}} \\ 
        & & \textbf{P} & \textbf{R} &\textbf{F1} & \textbf{P} & \textbf{R} & \textbf{F1} & \textbf{P} & \textbf{R} & \textbf{F1} \\ \hline
        \multirow{2}{*}{gpt-3.5-turbo} & baseline & 0.557 & 0.574 & 0.540 & 0.545 & 0.579 & 0.525 & 0.599 & 0.659 & 0.593 \\
        & improved & 0.581 & 0.597 & 0.563 & 0.576 & 0.609 & 0.554 & 0.625 & 0.684 & \textbf{0.618} \\ \hline
        \multirow{2}{*}{gpt-4} & baseline & 0.650 & 0.661 & 0.632 & 0.641 & 0.651 & 0.624 & 0.650 & 0.685 & 0.641 \\
        & improved & 0.682 & 0.689 & 0.661 & 0.678 & 0.683 & 0.657 & 0.676 & 0.709 & \textbf{0.665} \\
    \hline
    \end{tabular}
    \end{adjustbox}
\end{table}

From the lens of relations, as shown in the detailed results of GPT-4 (Table~\ref{tab:gpt-4-results}), LLMs perform well when the relation has a limited domain and/or range, for example, \textit{PersonHasNobelPrize}, \textit{CountryHasOfficialLanguage}, and \textit{CompoundHasParts}. On the other hand, LLMs perform poorly for relations such as \textit{PersonHasEmployer}, \textit{PersonHasProfession}, and \textit{PersonHasAutobiography}. This may be due to two reasons: firstly, LLMs have limited knowledge about public figures and their personal information (except for famous ones). Secondly, the unlimited answer space for such relations could increase the difficulty of prediction. The results show that LLMs perform relatively well on the knowledge of geography, as GPT-4 achieved F1-scores of 0.629 on \textit{CityLocatedAtRiver}, 0.763 on \textit{CountryBordersCountry}, 0.855 on \textit{RiverBasinsCountry}, and 0.581 on \textit{StateBordersState}, and the performance is inversely correlated with the size of the object range. The knowledge of public figures contained in LLMs could be an interesting topic to investigate since their performance across different aspects varies significantly. While LLMs correctly handle every instance of \textit{PersonHasNobelPrize}, they also demonstrate relatively strong performance in areas such as place of birth and death, cause of death, and spouses. However, their performance tends to be deficient when it comes to details about individuals' employers and professions.

\begin{table}[ht]
    \centering
    \caption{The results of probing GPT-4. For each relation, the improved disambiguation method used is listed, and the best F1-scores among the three settings are highlighted.}
    \label{tab:gpt-4-results}
    \begin{adjustbox}{width=1\textwidth}
    \begin{tabular}{l|ccc|ccc|ccc|c}
    \hline
        \multicolumn{1}{c|}{\multirow{2}{*}{\textbf{Relation}}} & \multicolumn{3}{c}{\textbf{question}} & \multicolumn{3}{c}{\textbf{triple}} & \multicolumn{3}{c|}{\textbf{context}} & \multicolumn{1}{c}{\multirow{2}{*}{\textbf{Disambiguation}}}\\
        & \textbf{P} & \textbf{R} & \textbf{F1} & \textbf{P} & \textbf{R} & \textbf{F1} & \textbf{P} & \textbf{R} & \textbf{F1} & \\ \hline
        BandHasMember & 0.576 & 0.632 & 0.573 & 0.591 & 0.627 & \textbf{0.581} & 0.510 & 0.627 & 0.527 & Keyword \\
        CityLocatedAtRiver & 0.780 & 0.562 & 0.615 & 0.775 & 0.578 & \textbf{0.629} & 0.648 & 0.504 & 0.533 & LM \\
        CompanyHasParentOrganisation & 0.590 & 0.755 & \textbf{0.590} & 0.560 & 0.745 & 0.563 & 0.512 & 0.810 & 0.520 & Baseline\\
        CompoundHasParts & 0.782 & 0.976 & 0.837 & 0.782 & 0.964 & 0.835 & 0.787 & 0.981 & \textbf{0.843} & Case \\
        CountryBordersCountry & 0.802 & 0.685 & 0.730 & 0.806 & 0.688 & 0.734 & 0.829 & 0.723 & \textbf{0.763} & Baseline \\
        CountryHasOfficialLanguage & 0.956 & 0.854 & 0.883 & 0.949 & 0.858 & 0.883 & 0.938 & 0.873 & \textbf{0.886} & Keyword \\
        CountryHasStates & 0.796 & 0.809 & 0.800 & 0.754 & 0.748 & 0.750 & 0.805 & 0.816 & \textbf{0.807} & LM \\
        FootballerPlaysPosition & 0.685 & 0.693 & 0.680 & 0.710 & 0.733 & \textbf{0.708} & 0.545 & 0.565 & 0.550 & Case \\
        PersonCauseOfDeath & 0.765 & 0.783 & 0.762 & 0.795 & 0.803 & 0.793 & 0.800 & 0.803 & \textbf{0.798} & Baseline \\
        PersonHasAutobiography & 0.478 & 0.471 & \textbf{0.461} & 0.458 & 0.486 & \textbf{0.461} & 0.475 & 0.471 & 0.459 & Keyword \\
        PersonHasEmployer & 0.362 & 0.343 & 0.327 & 0.353 & 0.357 & \textbf{0.328} & 0.325 & 0.397 & 0.321 & Case \\
        PersonHasNobelPrize & 1.000 & 1.000 & \textbf{1.000} & 1.000 & 1.000 & \textbf{1.000} & 1.000 & 1.000 & \textbf{1.000} & Baseline \\
        PersonHasNumberOfChildren & 0.550 & 0.550 & 0.550 & 0.520 & 0.520 & 0.520 & 0.690 & 0.690 & \textbf{0.690} & None \\
        PersonHasPlaceOfDeath & 0.670 & 0.730 & 0.670 & 0.690 & 0.730 & 0.690 & 0.783 & 0.810 & \textbf{0.785} & Baseline \\
        PersonHasProfession & 0.494 & 0.420 & 0.427 & 0.538 & 0.422 & \textbf{0.444} & 0.390 & 0.408 & 0.363 & Case \\
        PersonHasSpouse & 0.687 & 0.690 & 0.685 & 0.652 & 0.660 & 0.651 & 0.718 & 0.750 & \textbf{0.727} & LM \\
        PersonPlaysInstrument & 0.566 & 0.565 & 0.531 & 0.559 & 0.519 & 0.507 & 0.559 & 0.597 & \textbf{0.534} & Case \\
        PersonSpeaksLanguage & 0.747 & 0.813 & 0.744 & 0.755 & 0.836 & \textbf{0.759} & 0.757 & 0.808 & 0.742 & Baseline \\
        RiverBasinsCountry & 0.841 & 0.946 & \textbf{0.855} & 0.841 & 0.931 & 0.852 & 0.827 & 0.941 & 0.852 & Case \\
        SeriesHasNumberOfEpisodes & 0.590 & 0.590 & 0.590 & 0.530 & 0.530 & 0.530 & 0.690 & 0.690 & \textbf{0.690} & None \\
        StateBordersState & 0.608 & 0.600 & 0.567 & 0.619 & 0.608 & \textbf{0.581} & 0.612 & 0.618 & 0.578 & LM \\ 
    \hline
    \end{tabular}
    \end{adjustbox}
\end{table}

\subsection{Retrieval-Augmented Prediction}

Providing relevant corpus as context to LLMs is an established method for improving model performance~\cite{brown-et-al-2020-gpt-3}. As such, we experimented with various sources and forms of context and selected the best ones for each relation. In particular, we experimented with using the introduction paragraphs of the Wikipedia article for the subject entity, the Infobox of the Wikipedia article for the subject entity in JSON format, as well as relation-specific sources such as IMDb. The effect of providing context varies for different models. It is observed gpt-3.5-turbo benefits from the context more compared with GPT-4. Reflected from F1-scores, the retrieval-augmented context setting exhibits an improvement of 0.055 compared with the question prompting setting for gpt-3.5-turbo and 0.004 for GPT-4. 

In contrast to our intuition, adding context knowledge does not enhance the performance of GPT-4 in all relations as compared to only proving the few-shot examples, where only 10 out of 21 relations achieved better results in the context setting compared to the question and triple settings. Several factors may contribute to this, including the presence of a knowledge gap and misaligned entity representations between Wikipedia and Wikidata. These factors could impact model performance, particularly when LLMs heavily rely on context enriched from Wikipedia. An example is \textit{FootnallerPlaysPosition}, where we have noted discrepancies between Wikipedia and Wikidata in the names used to represent identical or similar positions on the field. The investigation of this knowledge gap is explained in Section~\ref{gap} and warrants further examination. 

For most relations, where augmented context improved the performance, the introduction and Infobox of the Wikipedia page are sufficient based on the performance and the cost balance. Notable exceptions to the above are the \textit{CountryHasState} and \textit{SeriesHasNumberOfEpisodes} relations, where we augmented relation-specific context. For the \textit{SeriesHasNumberOfEpisodes} relation, except for the previous two sources, we augmented the context from IMDb. The information on IMDb was added to the prompt prefaced by the label ``IMDb'', and the model was asked to use this information (if it was available) to provide an answer. Moreover, for the \textit{CountryHasState} relation, we discovered that GPT-4 would treat `state' more like the definition of `country' than that of the administrative division entity. Therefore, we experimented with providing the model with ``Administrative Division of [entity]'' Wikipedia page content, which outperformed the question setting for 0.007 of the F1-score.

\subsection{Disambiguation}

When using the baseline disambiguation method, we observed disambiguation mistakes in 13 relations. These errors are categorized into two groups: \textbf{surface} disambiguation errors, in which the model produced the same strings of entities as the ground truths but assigned incorrect QIDs, and \textbf{deep} disambiguation errors, where the model associated the same entities with different names (i.e., aliases) and also assigned incorrect QIDs. In this study, we focus only on addressing the former category while reserving discussion of the latter for future research. To tackle this challenge, we implemented improved disambiguation methods with the dual objective of rectifying errors to the fullest extent possible and concurrently reducing computational complexity.

From Table~\ref{tab:model-performance}, we can observe an average increase in F1-scores of 0.0256 for all settings in the case of gpt-3.5-turbo and 0.0289 for GPT-4. For the 13 relations where improved disambiguation methods are applied, Table~\ref{tab:gpt-4-results} listed the best-performing disambiguation method for each relation. Notably, for 3 relations (\textit{CompoundHasParts}, \textit{PersonPlaysInstrument}, and \textit{RiverBasinsCountry}), the issues have been successfully solved. However, the rest 8 relations still remain either 2 or fewer unsolved errors, and 2 relations (\textit{BandHasMember} and \textit{StateBordersState}) face more than 7 unsolved errors, exceeding the capacity of their respective methods.

Given that the \textit{wbsearchentities} Action API relies on label and alias-based searching, there's a potential issue when LLMs predict objects with labels that are absent from the label and aliases of the corresponding Wikidata entity. This mismatch can lead to an incomplete list of candidate entities. From this perspective, LLMs have the ability to contribute to knowledge engineering by enriching the labels and aliases associated with Wikidata entities.

\section{Discussion}\label{discussion}
\subsection{Wikidata Quality}

During the development of our pipeline and the evaluation of the results, it became apparent that the quality of Wikidata is an important issue, a problem that has also been discussed in previous works~\cite{piscopo-simperl-2019,shenoy-et-al-2022}. For example, a large number of elements are missing for the relation \textit{CompoundHasParts}, and many objects violate the value-type constraint of properties. In this situation, our proposed method would be useful for automatically providing suggestions and candidates for incomplete triples and thus enriching Wikidata by improving its quality. Moreover, it is possible to use LLMs to align the knowledge contained in Wikidata with the knowledge contained in Wikipedia and complete the triples of Wikidata using the Wikipedia articles as context. Furthermore, the performance of the LLMs on the object prediction task can be used as a metric to gauge the completeness of Wikidata entities. In cases where the difference between the predictions of the LLMs and the ground truth is substantial, the entity can be suggested to Wikidata editors for review using a recommender system, such as the one described by~\cite{alghamdi-et-al-2021}. Finally, the labels (synonyms) of Wikidata entities are incomplete, which limits the ability of our disambiguation method since the system that retrieves the candidate entities needs labels and aliases to match the given string. 

\subsection{Knowledge Gap}\label{gap}

Through our efforts to use Wikipedia as relevant context to improve the performance of LLMs in the object prediction task, we observed a significant knowledge gap between Wikipedia and Wikidata, which caused the performance of the model to deteriorate when provided with context sourced from Wikipedia for some of the relations. To elucidate the cause of this phenomenon, we manually inspected several of these instances and realized that the information contained in Wikidata is different from the information contained in Wikipedia. One such example is the subject-relation pair \textit{Ferrari S.p.A., CompanyHasParentOrganisation}, for which LLMs correctly predicted the object \textit{Exor}, matching the information on Wikipedia and the official report from Ferrari in 2021, whereas Wikidata contains the object \textit{Ferrari N.V.}, which is outdated. This knowledge gap between Wikipedia and Wikidata is an open issue, and LLMs, either alone or by supporting human editors and suggesting edits, could play a pivotal role in addressing this issue and improving the data quality and recency of information contained in Wikidata. Finally, the knowledge gap is not limited to Wikidata and Wikipedia but appears to exist between LLMs as well. Specifically, as seen in Table~\ref{tab:online-evaluation}, gpt-3.5-turbo outperforms the larger GPT-4 in two of the relations. Based on this, it stands to reason that different LLMs can contain different knowledge, and therefore, using an ensemble of LLMs with complementary strengths can lead to an improvement in performance.   

\section{Conclusion}\label{conclusion}
Within the scope of the ISWC 2023 LM-KBC challenge, this work aimed at developing a method to probe LLMs for predicting the objects of Wikidata triples given the subject and relation. Our best-performing method achieved state-of-the-art results with a macro-averaged F1-score of 0.7007 across all relations, with GPT-4 having the best performance on the \textit{PersonHasNobelPrize} relation and achieving a score of 1.0, while only achieving a score of 0.328 on the \textit{PersonHasEmployer} relation. These results show that LLMs can be effectively used to complete knowledge bases when used in the appropriate context. At the same time, it is important to note that, largely due to the gaps in their knowledge, fully automatic knowledge engineering using LLMs is not currently possible for all domains, and a human-in-the-loop is still required to ensure the accuracy of the information.

\begin{acknowledgments}
This work was partly funded by the HE project MuseIT, which has been co-founded by the European Union under the Grant Agreement No 101061441. Views and opinions expressed are, however, those of the authors and do not necessarily reflect those of the European Union or European Research Executive Agency.
\end{acknowledgments}

\bibliography{ms}

\newpage
\appendix

\section{Online Evaluation Results}\label{online}
\begin{table}[h!]
    \centering
    \caption{The online evaluation results from CodaLab. The results are aggregated from the highest ones in the three settings for each relation and model. The online evaluation results from CodaLab. The results are aggregated from the highest ones in the three settings for each relation and model. The outcomes conducted in online evaluation have demonstrated superior results compared to the offline ones, with particular significance observed in the cases of \textit{CompoundHasParts} and \textit{CityLocatedAtRiver}. This phenomenon can be attributed to the revision of online ground truths. Additionally, this observation emphasizes that LLMs have the potential to enhance the overall quality of Wikidata.}
    \label{tab:online-evaluation}
    \begin{adjustbox}{width=1\textwidth}
    \begin{tabular}{l|cccc|cccc}
        \hline
        \multicolumn{1}{c|}{\multirow{2}{*}{\textbf{Relation}}} & \multicolumn{4}{c|}{\textbf{gpt-3.5-turbo}} & \multicolumn{4}{c}{\textbf{GPT-4}} \\
        & \textbf{P} & \textbf{R} & \textbf{F1} & \textbf{Setting} & \textbf{P} & \textbf{R} & \textbf{F1} & \textbf{Setting} \\ \hline
        BandHasMember & 0.5378 & 0.5830 & 0.5295 & triple & 0.5905 & 0.6331 & \textbf{0.5838} & triple \\
        CityLocatedAtRiver & 0.5500 & 0.4723 & 0.4845 & context & 0.7600 & 0.6538 & \textbf{0.6792} & triple \\
        CompanyHasParentOrganisation & 0.4300 & 0.7500 & 0.4267 & context & 0.6100 & 0.7650 & \textbf{0.6100} & question \\
        CompoundHasParts & 0.9591 & 0.9659 & 0.9615 & context & 0.9962 & 1.0000 & \textbf{0.9978} & context \\
        CountryBordersCountry & 0.8628 & 0.7756 & \textbf{0.8107} & context & 0.8292 & 0.7699 & 0.7937 & context \\
        CountryHasOfficialLanguage & 0.9313 & 0.8731 & 0.8814 & question & 0.9379 & 0.8821 & \textbf{0.8932} & context \\
        CountryHasStates & 0.7926 & 0.7772 & 0.7823 & context & 0.8048 & 0.8156 & \textbf{0.8073} & context \\
        FootballerPlaysPosition & 0.6400 & 0.6333 & 0.6323 & triple & 0.7100 & 0.7333 & \textbf{0.7083} & triple \\
        PersonCauseOfDeath & 0.7600 & 0.7833 & 0.7550 & question & 0.8000 & 0.8033 & \textbf{0.7983} & context \\
        PersonHasAutobiography & 0.4337 & 0.5000 & 0.4490 & context & 0.4483 & 0.4850 & \textbf{0.4583} & context \\
        PersonHasEmployer & 0.3053 & 0.4087 & 0.3134 & context & 0.3533 & 0.3567 & \textbf{0.3282} & triple \\
        PersonHasNoblePrize & 0.9900 & 0.9900 & 0.9900 & question & 1.0000 & 1.0000 & \textbf{1.0000} & question \\
        PersonHasNumberOfChildren & 0.6900 & 0.6900 & 0.6900 & context & 0.7000 & 0.7000 & \textbf{0.7000} & context \\
        PersonHasPlaceOfDeath & 0.6150 & 0.7800 & 0.6167 & context & 0.7833 & 0.8100 & \textbf{0.7850} & context \\
        PersonHasProfession & 0.2875 & 0.3927 & 0.3029 & context & 0.5375 & 0.4159 & \textbf{0.4395} & triple \\
        PersonHasSpouse & 0.7583 & 0.7850 & \textbf{0.7650} & context & 0.7083 & 0.7450 & 0.7183 & context \\
        PersonPlaysInstrument & 0.3987 & 0.4946 & 0.4087 & context & 0.5485 & 0.5924 & \textbf{0.5279} & context \\
        PersonSpeaksLanguage & 0.8683 & 0.6893 & 0.7344 & triple & 0.7550 & 0.8360 & \textbf{0.7589} & triple \\
        RiverBasinsCountry & 0.7869 & 0.8986 & 0.8054 & context & 0.8408 & 0.9463 & \textbf{0.8549} & question \\
        SeriesHasNumberOfEpisodes & 0.6200 & 0.6300 & 0.6233 & context & 0.6900 & 0.6900 & \textbf{0.6900} & context \\
        StateBordersState & 0.5753 & 0.5898 & 0.5435 & context & 0.6139 & 0.6135 & \textbf{0.5811} & triple \\ \hline
        \multicolumn{1}{c|}{Zero-object cases} & 0.4708 & 0.7559 & 0.5802 & / & 0.5026 & 0.9202 & \textbf{0.6501} & / \\ \hline
        \multicolumn{1}{c|}{Average} & 0.6568 & 0.6887 & 0.6432 & / & 0.7151 & 0.7260 & \textbf{0.7007} & / \\
        \hline
    \end{tabular}
    \end{adjustbox}
\end{table}

\section{Prompt Templates}\label{prompt}
\noindent \textbf{BandHasMember} \\
Who are the members of \{subject\_entity\}? Format the response as a Python list such as ["answer\_a", "answer\_b"]. \\

\noindent \textbf{CityLocatedAtRiver} \\
Which river is \{subject\_entity\} located at? Format the response as a Python list such as ["answer\_a", "answer\_b"]. \\ 

\noindent \textbf{CompanyHasParentOrganisation} \\
\{subject\_entity\} is a subsidiary of which company? Return a Python list with an empty string (i.e. [""]) if none. Format the response as a Python list such as ["answer\_a", "answer\_b"]. \\

\noindent \textbf{CountryBordersCountry} \\
Which countries share borders with \{subject\_entity\}? Format the response as a Python list such as ["answer\_a", "answer\_b"]. \\

\noindent \textbf{CountryHasOfficialLanguage} \\
What is the official language of \{subject\_entity\}? Format the response as a Python list such as ["answer\_a", "answer\_b"]. \\

\noindent \textbf{CountryHasStates} \\
What are the first-level administrative territorial entities of \{subject\_entity\}? Format the response as a Python list such as ["answer\_a", "answer\_b"]. \\

\noindent \textbf{FootballerPlaysPosition} \\
What position does \{subject\_entity\} play in football? Format the response as a Python list such as ["answer\_a", "answer\_b"]. \\

\noindent \textbf{PersonCauseOfDeath} \\
What caused the death of \{subject\_entity\}? If none or still alive, return [""]. Format the response as a Python list such as ["answer\_a", "answer\_b"]. \\

\noindent \textbf{PersonHasAutobiography} \\What is the title of \{subject\_entity\}'s autobiography? Format the response as a Python list such as ["answer\_a", "answer\_b"]. \\

\noindent \textbf{PersonHasEmployer} \\
Who is \{subject\_entity\}'s employer? Format the response as a Python list such as ["answer\_a", "answer\_b"]. \\

\noindent \textbf{PersonHasNoblePrize} \\
Which Nobel Prize did \{subject\_entity\} receive? Select from this list: ["Nobel Peace Prize", "Nobel Prize in Literature", "Nobel Prize in Physics", "Nobel Prize in Chemistry", "Nobel Prize in Physiology or Medicine"]. Return a Python list with an empty string (i.e. [""]) if none. Format the response as a Python list such as ["answer\_a", "answer\_b"]. \\

\noindent \textbf{PersonHasNumberOfChildren} \\
How many children does \{subject\_entity\} have? Return the string format of the number only. Format the response as a Python list such as ["answer\_a", "answer\_b"]. \\

\noindent \textbf{PersonHasPlaceOfDeath} \\
Where did \{subject\_entity\} die? Return a Python list with an empty string (i.e. [""]) if he or she is still alive. Format the response as a Python list such as ["answer\_a", "answer\_b"]. \\

\noindent \textbf{PersonHasProfession} \\
What is \{subject\_entity\}'s profession or occupation? Format the response as a Python list such as ["answer\_a", "answer\_b"]. \\

\noindent \textbf{PersonHasSpouse} \\
What is the name of the spouse of \{subject\_entity\}? Format the response as a Python list such as ["answer\_a", "answer\_b"]. \\

\noindent \textbf{PersonPlaysInstrument} \\
What instruments does \{subject\_entity\} play? Format the response as a Python list such as ["answer\_a", "answer\_b"]. \\

\noindent \textbf{PersonSpeaksLanguage} \\
What languages does \{subject\_entity\} speak? Format the response as a Python list such as ["answer\_a", "answer\_b"]. \\

\noindent \textbf{RiverBasinsCountry} \\
In which country can you find the \{subject\_entity\} river basin? Format the response as a Python list such as ["answer\_a", "answer\_b"]. \\

\noindent \textbf{SeriesHasNumberOfEpisodes} \\
How many episodes does the series \{subject\_entity\} have? Return the string format of the number. Format the response as a Python list such as ["answer\_a", "answer\_b"]. \\

\noindent \textbf{CompoundHasParts} \\
What are the chemical components of \{subject\_entity\}? Return the full name of components such as ["carbon", "nitrogen"]. Format the response as a Python list such as ["answer\_a", "answer\_b"]. \\

\noindent \textbf{StateBordersState} \\
Which states border the state of \{subject\_entity\}? Format the response as a Python list such as ["answer\_a", "answer\_b"].

\end{document}